\documentclass[compactaffiliation]{Interspeech}

\interspeechcameraready

\title{Clinical Annotations for Automatic Stuttering Severity Assessment}

\author[affiliation={1,2}]{Ana Rita}{Valente}
\author[affiliation={3}]{Rufael}{Marew}
\author[affiliation={3}]{Hawau Olamide}{Toyin}
\author[affiliation={3}]{Hamdan}{Al-Ali}
\author[affiliation={4}]{Anelise}{Bohnen}
\author[affiliation={2}]{Inma}{Becerra}
\author{Elsa Marta}{Soares}
\author[affiliation={2}]{Gonçalo}{Leal}
\author[affiliation={3}]{Hanan}{Aldarmaki}

\affiliation{}{University of Aveiro}{Portugal}
\affiliation{}{SpeechCare Center}{}
\affiliation{}{MBZUAI}{UAE}
\affiliation{}{Instituto Brasileiro de Fluência}{Brazil}
\email{
rita.valente@ua.pt, 
hanan.aldarmaki@mbzuai.ac.ae 
}

\keywords{stuttering assessment, disfluency detection}

\usepackage{multirow}
\usepackage[textsize=tiny]{todonotes}
\newcommand{\ra}[1]{\renewcommand{\arraystretch}{#1}}
\usepackage{pifont}

\usepackage{comment}

\begin{document}


\maketitle

\begin{abstract}
    
    Stuttering is a complex disorder that requires specialized expertise for effective assessment and treatment. This paper presents an effort to enhance the FluencyBank dataset with a new stuttering annotation scheme based on established clinical standards. To achieve high-quality annotations, we hired expert clinicians to label the data, ensuring that the resulting annotations mirror real-world clinical expertise. The annotations are multi-modal, incorporating audiovisual features for the detection and classification of stuttering moments, secondary behaviors, and tension scores.   In addition to individual annotations, we additionally provide a test set with highly reliable annotations based on expert consensus for assessing individual annotators and machine learning models. Our experiments and analysis illustrate the complexity of this task that necessitates extensive clinical expertise for valid training and evaluation of stuttering assessment models. 
    
\end{abstract}


\section{Introduction}

Stuttering is a neurodevelopmental and multidimensional communication disorder that begins early in speech and language development \cite{manning2023,
bloodstein2021, yairi2013epidemiology}. It is characterized by involuntary disruptions in speech fluency, such as repetitions, prolongations, and blocks, which are inconsistent and variable \cite{bloodstein2021,yaruss2007,
guitar2019}. 
In addition to speech disruptions, stuttering also includes various behavioral, emotional, and cognitive components. Behaviorally, stuttering is marked by speech interruptions, muscle tension, and other behaviors related to speech difficulties (e.g. escape behaviors). These speech production challenges can trigger emotional responses such as fear, guilt, shame, or anxiety related to speaking and anticipation of stuttering.  

Stuttering can negatively affect several areas of daily life for both children and adults, who often report feelings of ``loss of control" \cite{blood2016,tichenor2022}. 
The affective, behavioral, and cognitive aspects of stuttering can limit their interactions in personal, social, and professional contexts \cite{craig2009,yaruss2004}. Several cognitive factors play a role in stuttering, such as avoidance behaviors, intended to alleviate emotional and behavioral stress \cite{bloodstein2021}. 
Understanding the multiplicity of stuttering disorder involves the integration of all the components mentioned above in a comprehensive assessment process \cite{yaruss2007,coleman2014}. Specifically, with regard to the behavioral component, it is crucial to identify the types of disfluencies and to determine their frequency, duration, and the tension with which they are produced. Accurately characterizing these disfluencies is essential for determining the severity of stuttering, establishing treatment goals, tracking progress throughout the intervention, and monitoring post-treatment results \cite{guitar2019}. However, the variable nature of this fluency disorder complicates the assessment process. Stuttering severity fluctuates across different speaking situations, with various interlocutors, audience sizes, or conversation topics \cite{manning2023, 
guitar2019,yairi2023}.

Traditionally, the assessment of stuttering moments in Adults Who Stutter (AWS) is conducted using audio-visual recordings of communicative situations, typically in a therapeutic setting \cite{valente2018,riley2009}. The analysis is based on the evaluators' perceptions and the individuals' self-reports of difficulties. Moreover, it is still predominantly performed manually or with limited use of technology, making it challenging, time-consuming, and reducing its reliability and validity \cite{valente2015}. Several commendable efforts to advance automatic processing methods for stuttered speech exist, including the FluencyBank dataset \cite{fluencybank} which includes annotated videos of adults who stutter. However, annotations of these samples include only the primary disfluency type, which is insufficient for clinically-accurate severity assessment. Furthermore, existing annotation efforts rely on non-expert listeners \cite{sep28k} which results in unreliable annotations. In this work, we propose a comprehensive multi-dimensional annotation scheme for stuttering severity assessment that aligns with clinical practice. We hire expert Speech-Language Pathologists (SLPs) to perform the annotations on the public segments of the FluencyBank dataset. We analyze inter-annotator agreement, and provide consenus-based annotations for the test set. The resulting annotations can be used for training and validating multi-modal, multi-label automatic stuttering assessment systems. The annotations will be made publicly available to align stuttering research efforts with clinical use cases. 

\section{Related Work}

Research on automatic stuttering detection, also called disfluency detection, has been facilitated by several publicly available data sets that include audio or video recordings of adults who stutter. 
~\textbf{FluencyBank} ~\cite{fluencybank} is a database that includes audio/video recordings from various participants, including children and adults who stutter (C/AWS). FluencyBank is a component of the broader TalkBank project, which serves as a comprehensive resource for studying fluency disorders.
The stuttering data is associated with text transcripts with disfluency markers, and has been influential in facilitating research in automatic stuttering detection. However, inaccuracies in the temporal alignment of transcriptions and disfluency annotations have been noted ~\cite{riad-etal-2020-identification, sep28k}. 
The ~\textbf{Stuttering Events in Podcasts (SEP-28k)} dataset ~\cite{sep28k} contains over 28,000 clips labeled with five stuttering types: blocks, prolongations, sound repetitions, word repetitions, and interjections. The audio, sourced from public podcasts, features interviews with and by people who stutter. 
The data set was divided into 3-second audio segments, which are then annotated by nonclinical annotators, and the same annotation methodology was applied to the FluencyBank interview section. From a clinical perspective, these annotations are not suitable for stuttering assessment as they rely solely on short segments of audio by annotators who lack clinical expertise.
~\textbf{UCLASS} ~\cite{uclass} provides audio recordings of persons who stutter along with orthographic transcriptions with time markers, and stuttering types were annotated by speech-language pathologists. While these annotations are more reliable that previous efforts, they lack the additional visual dimensions necessary for accurate assessment.  

Complementary efforts include the ~\textbf{LibriStutter} dataset ~\cite{libriStutter}, which is an artificially generated dataset derived from the public LibriSpeech ASR corpus, and includes 5 stuttering types. 
~\textbf{Kassel State of Fluency (KSoF)} ~\cite{ksof} is a therapy-based dataset designed to monitor speech behavior over time, particularly for individuals who have undergone stuttering therapy. It includes over 5,500 clips of German speech labeled with six stuttering-related event types, including a unique category for speech modifications specific to therapy.

While such resources are valuable for building robust automatic speech recognition systems that are inclusive of persons who stutter, they are limited in their clinical applicability. As described in the introduction, stuttering severity assessment in a therapeutic setting is a complex, multi-dimensional process that involves the identification of behavioral elements beyond the auditory; none of the existing annotations for stuttering include the full-range of clinical annotations needed for accurate stuttering severity assessment.

\section{Annotation Methodology}
\label{sec:guidelines}

Stuttering moments, defined as interruptions in speech that represent the observable aspect of the stuttering disorder, can be classified based on frequency, duration, disfluency type, and the type/severity of secondary behaviors \cite{
bloodstein2021}. The combination of these quantitative behavioral measures (i.e., frequency, duration, and severity of secondary behaviors) is used to determine stuttering severity \cite{
guitar2019, riley2009}.
Frequency is typically expressed as the percentage of stuttered syllables, calculated by determining the total number of syllables produced by the speaker and identifying those in which stuttering occurs \cite{guitar2019}. Duration refers to the amount of time a stuttering interruption lasts \cite{
gillam2009}. Various classification systems exist for disfluency types, such as Stuttering-Like Disfluencies (SLD) \cite{yairi1992} and the Lidcombe Behavioral Data Language (LBDL) \cite{
packman2000}. The LBDL taxonomy has demonstrated higher inter-judge agreement in 10-second samples \cite{teesson2003}.

Stuttering moments are often accompanied by associated or secondary behaviors, which vary in severity and indicate the individual's reactions and coping strategies for fluency disruptions \cite{
yairi2023}. Similar to the classification of disfluency types, several systems have been developed to categorize secondary behaviors. Examples include the LBDL taxonomy \cite{
packman2000} and classifications used in specific stuttering assessment tools, such as the Stuttering Severity Instrument \cite{riley2009}. In this work, we combine three types of clinical classification systems for disfluencies, secondary behaviors, and tension, to enable comprehensive stuttering severity assessment. Our annotation scheme, designed in collaboration with expert SLPs, is described below. \\

\subsection{Annotation Process and Guidelines}

\textbf{Annotators:} Three speech and language pathologists, with experience ranging from 2 to 40 years, independently annotated and analyzed the audiovisual samples.
All annotators had professional backgrounds spanning both clinical practice and research/teaching, including work with Adults Who Stutter (AWS). 
\noindent\textbf{Annotator Training:} To ensure consistency in the analysis of stuttering moments, the annotators were trained using a step-by-step manual outlining the guidelines for annotation, leveraging the EUDICO Linguistic Annotator (ELAN), which facilitates the creation, editing, searching, and visualization of video and audio annotations \cite{wittenburg2006}. This tool enabled the creation of multimodal annotations by integrating video, acoustic waveforms, and speech into a unified workspace. 
A designated tier labeled ``Stuttering moments" was specifically created for annotating instances of stuttering, while a pre-existing tier displaying orthographic transcription from FluencyBank provided additional support during the annotation process.

 The annotation manual was based on the reliable and valid severity assessment instrument developed by Valente \cite{valente2018}. It included detailed screenshots illustrating the use of ELAN for audiovisual sample assessment, guiding the annotators in identifying and classifying each stuttering moment in terms of disfluency type, secondary behaviors, and tension level.

For \textbf{boundary Annotations}, the annotators marked the onset and offset of each stuttering instance by dragging the mouse from the beginning to the end of the event, with the duration automatically calculated. To improve accuracy, the visualization of acoustic waveforms was utilized, following the recommendations of Gillam et al. \cite{gillam2009}. Specifically: (1) for repetitions, the stuttering onset is the point where the repeated speech begins, and the offset is where fluent speech resumes; (2) for prolongations or blocks, the onset is the start of the disfluent speech sound, with the offset marked by the next fluent sound.

After identifying each stuttering moment, the annotators classified the \textbf{disfluency type} based on the LBDL taxonomy, secondary behaviors using Riley’s (2009) SSI-4 classification, and the tension level according to a 0-3 scale, as proposed by Boey et al. \cite{boey2007}.
The LBDL taxonomy categorizes disfluencies into three broad groups, with seven specific descriptors. In this study, the following LBDL descriptors were used: (1) \textbf{SR}: Syllable Repetition, indicating repeated movement of an entire syllable; (2) \textbf{ISR:} Incomplete Syllable Repetition, repetition of parts of syllables; (3) \textbf{MUR:} Multisyllable Unit Repetition, repetitions involving multiple syllables; (4) \textbf{P:} sound Prolongation; and (5) \textbf{B: } Blocks. 

\textbf{Secondary behaviors} were classified into four types: (1) \textbf{V: } Verbal behaviors, including noisy breathing, whistling, sniffing, blowing, or clicking sounds; (2) \textbf{FG:} Facial Grimaces, such as jaw jerking, tongue protrusion, lip pressing, jaw muscle tension, or eye movements; (3) \textbf{HM:} Head Movements, including backward, forward, or turning motions; and (4) \textbf{ME: } Movements of Extremities, such as arm and hand gestures, torso movements, or leg motions.

\textbf{Tension levels} were classified according to the scale proposed by Boey et al. \cite{boey2007}, which offers ease of use: 0 – no tension; 1 – mild, subtle physical tension; 2 – moderate tension, noticeable and distracting; and 3 – severe tension, highly distracting and produced with considerable effort.

\begin{table}
    \centering
     \caption{Mutli-dimensional Annotation Categories}
     \scalebox{0.8}{
    \begin{tabular}{ll}
        \toprule
        \multicolumn{2}{l}{\textbf{Disfluency Types (Primary)}}  \\
        \midrule
        SR  & Syllable Repetition\\
        ISR & Incomplete Syllable Repetition \\
        MUR & Multisyllable Unit Repetition \\
        P   & Sound Prolongation \\
        B   & Blocks \\
        \midrule
        \multicolumn{2}{l}{\textbf{Secondary Behaviors}}  \\
        \midrule
        V   & Verbal behaviors \\
        FG  & Facial Grimaces \\
        HM  & Head Movements\\
        ME  & Movements of Extremities  \\
        \midrule
        \multicolumn{2}{l}{\textbf{Tension Levels }} \\
        \midrule
        0   & No tension \\
        1   & Mild \\
        2   & Moderate  \\
        3   & Severe \\
        \bottomrule
    \end{tabular}
   }
    \label{tab:annotations}
\end{table}
\section{Analysis of Annotated Data}

\subsection{Data Sources }
All audiovisual samples were sourced from the FluencyBank database. In the audiovisual recordings, participants were positioned facing the camera, allowing for clear visualization of both the face and the shoulder girdle. 
\textbf{Reading:} A total of 30 audiovisual reading samples from Adults Who Stutter (AWS), with a cumulative duration of 1 hour and 10 minutes (mean duration of 2 minutes and 11 seconds per sample), were annotated. The reading material was an excerpt from \textit{In Pursuit of the Perfect Ham in Little-Known Friuli} by Russ Parsons, available through the SSI-4 instrument. 
\textbf{Interview:} 
36 audiovisual interview samples were analyzed, with a total duration of 6 hours and 15 minutes and a mean duration of 10 minutes and 41 seconds. 
According to Ratner and MacWhinney \cite{fluencybank}, interview questions were selected to maximize discussion of behavioral, affective and cognitive components of
chronic stuttering.

\subsection{Inter-Annotator Agreement}\label{sec:agreements}
\begin{figure}
    \centering
    \includegraphics[width=\columnwidth]{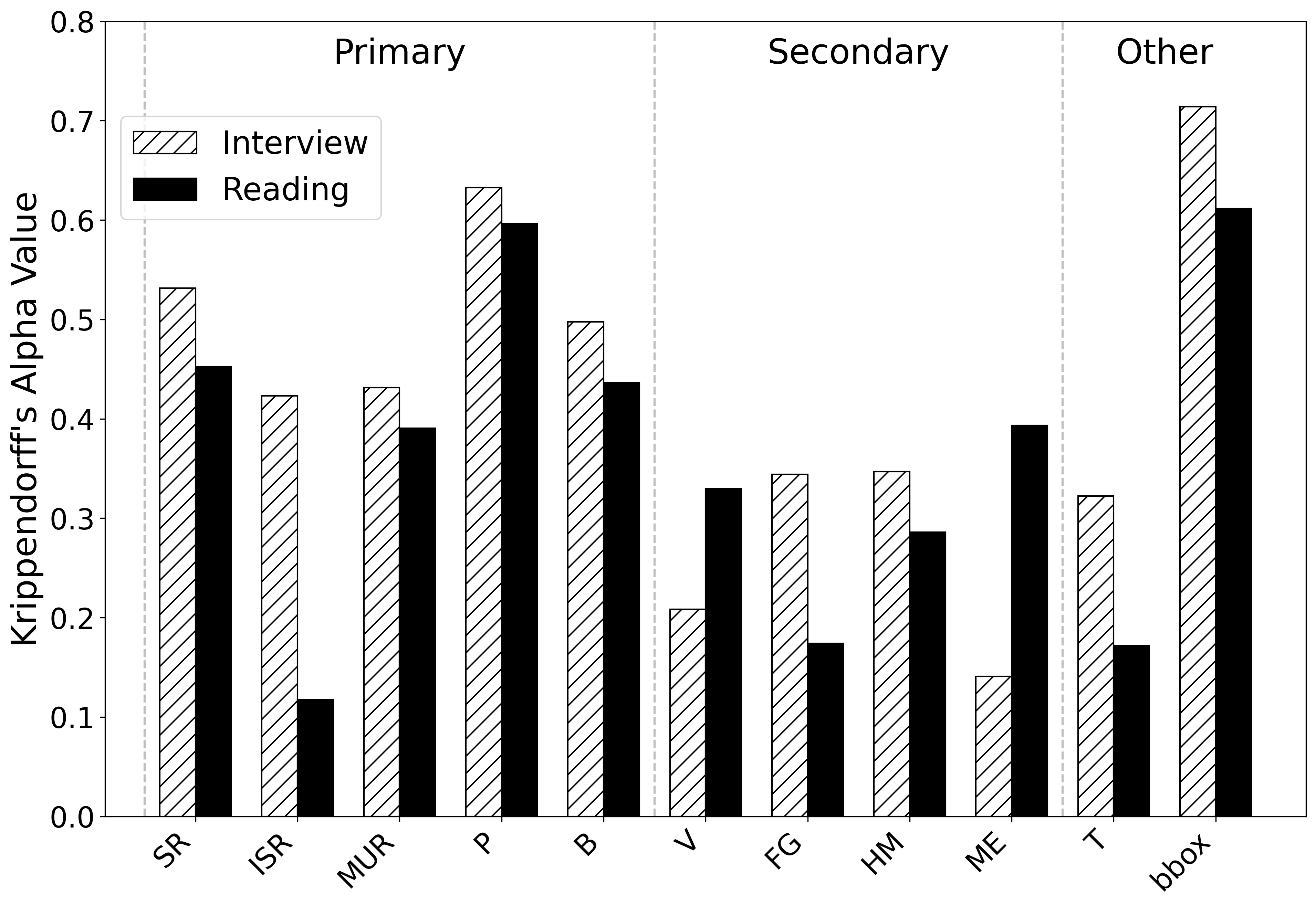}
    \caption{Krippendorff's alpha value for each stuttering type.
    }
    \label{fig:agreement}
\end{figure}

In total, we counted 1654 and 4037 non-overlapping stuttering spans in the reading and interview sections, respectively. Each span was annotated with a tension score, a primary type, and a secondary type. In some cases, only primary or secondary types were observed. 
We report the Inter-Annotator Agreement (IAA) scores for each type of annotation: time spans, primary and secondary types, and tension. For spans, we follow the guidelines described in \cite{braylan_measuring_2022}. First the annotations are grouped into similar time ranges using agglomerative clustering for fine-grained comparison. We used the intersection over Union (IoU) as a distance metric and calculated Krippendorff's $\alpha$, $KS$, and $\sigma$ as 0.68, 0.99, and 0.99, respectively. 
For the primary and secondary types, we used a binary distance metric for each class to calculate Krippendorff’s $\alpha$; the results are shown in figure \ref{fig:agreement}. For Tension, which is an ordinal variable, we used normalized rank Euclidean distance metric and got $\alpha= 0.18$. $KS=0.38$, and $\sigma=0.34$. 
Figure \ref{fig:agreement} also shows that the annotators agreed on the primary disfluency types more than the secondary behaviors.

\subsection{Test Set with Consensus Annotations}\label{sec:testset}

Fine-grained stuttering annotation exhibits high variability, even among expert SLPs, as shown in the previous section. 
To create a reliable test set, we conducted another round of annotations to resolve disagreements.  
Reading and interview samples from four AWS each, representing both genders and varying levels of severity, were selected based on instances where the two most experienced annotators showed the highest number of disagreements in classifying disfluency types.  
Together, these files constitute our test set. 
To facilitate the consensus annotation process, a combined ELAN file of each AWS was used, which included each annotator's individual annotations as well as a designated ``disagreement" tier. This tier highlighted the specific instances where discrepancies occurred among the annotators, thus directing the focus of the discussion. Instances where two or all three annotators were in agreement were excluded from the review. 
Concerning the reading samples, a total of 102 disagreements were reviewed and discussed during an online meeting that lasted four hours, involving all three annotators. For the interview samples, a total of 303 disagreements were discussed in two online meetings that lasted, in total, 6 hours.
Each disagreement was resolved through multiple visualizations of the stuttering event and in-depth discussions until consensus was reached on the final `gold standard' annotations. Counts of various stuttering and secondary behavior types in the resulting test set are shown in Table \ref{tab:test_stats}.

\begin{table}[h]
\caption{Counts of disfluency types and secondary behaviors as annotated in the `gold standard' test labels.}
    \centering
    \scalebox{0.8}{
    \begin{tabular}{|l|c|*{5}{c|}}
    \hline
    \multirow{2}{*}{\textbf{Type}} & \multirow{2}{*}{\textbf{Total}} & \multicolumn{5}{c|}{\textbf{Secondary Behaviour}} \\
    \cline{3-7}
    & & \textbf{V} & \textbf{FG} & \textbf{HM} & \textbf{ME} & \textbf{None} \\
    \hline
SR & 190 & 23 & 114 & 38 & 1 & 53 \\
\hline
ISR & 143 & 29 & 106 & 25 & 0 & 23 \\
\hline
MUR & 94 & 10 & 64 & 16 & 0 & 21 \\
\hline
P & 93 & 2 & 62 & 34 & 0 & 18 \\
\hline
B & 265 & 83 & 202 & 75 & 1 & 21 \\
\hline
None & 25 & 17 & 15 & 0 & 0 & 0 \\
\hline
\textbf{Total} & 732 & 164 & 563 & 188 & 2 & 136 \\
\hline
\end{tabular}
}
\label{tab:test_stats}
\end{table}

\subsection{Reported Challenges: }


The annotators reported several challenges: First, the task was inherently subjective, relying heavily on the evaluator’s perception, making it time-consuming and highly dependent on the evaluator’s experience. Second, many of the speech samples in the test test were produced by non-native English speakers, making it difficult to distinguish between disfluencies stemming from a fluency disorder and those resulting from limited proficiency in English. Third, some individuals appeared to have comorbidities (e.g., movement-related disorders), which further complicated decisions regarding the presence of secondary behaviors. Due to the lack of detailed information on additional diagnoses, we opted to include these cases in our annotations. Lastly, the annotation and analysis were further hindered by the poor quality of certain speech samples, including low-resolution of video and audio and the presence of background noise.
Overall, we see higher IAA scores in the interview section which was annotated after the annotation and discussion of the reading section.  This may indicate that annotator agreement could potentially improve as more and more data are annotated and discussed and they mutually reach better understanding of the task. Future annotation efforts could benefit from such smaller consecutive rounds of annotation and discussion by the same annotators. 
\subsection{Aggregation}
\label{sec:dataprep}

In this section, we compare the performance of each annotator, in addition to various methods to aggregate the annotations against the gold consensus labels. 
 We tested various annotation aggregation methods described in \cite{braylan_general_2023}, such as Best Available User (\textbf{BAU}), which computes the average deviation of each user compared to the other annotators over the entire dataset and selects the best user; Smallest Average Distance (\textbf{SAD}), which selects the annotation per item with the smallest average distance compared to other annotations; and Multi-Dimensional Annotation Scaling (called a MAS), which selects the most optimal annotation by comparing both annotator accuracy and per-item statistics. In addition, we tested a majority aggregation strategy (MAJ), selecting annotations that were provided by at least two annotators after clustering similar time spans as described in section \ref{sec:agreements}. 

 Table \ref{tab:segment_f1} shows the performance of individual annotators and these aggregation strategies against the gold labels. The evaluations were done using the segment-based evaluation of \textit{sed-eval}\footnote{\url{https://tut-arg.github.io/sed_eval/}} framework as described by \cite{app6060162}.

\begin{table}
\centering
\caption{F1 score for each annotator and aggregation method measured against gold labels across the classes using segment based evaluation as described in \cite{app6060162}}
\ra{1.2}
\resizebox{\columnwidth}{!}{
\begin{tabular}{@{}lccccccccc@{}}
\toprule
  & SR & ISR & MUR & P & B & V & FG & HM & Macro \\
\midrule
A1 & 0.61 & \textbf{0.64} & 0.53 & 0.66 & 0.60 & \textbf{0.65} & 0.80 & 0.68 & 0.76 \\
A2 & \textbf{0.69} & 0.44 & \textbf{0.73} & \textbf{0.70} & \textbf{0.74} & 0.64 & 0.81 & \textbf{0.69} & \textbf{0.79} \\
A3 & 0.57 & 0.54 & 0.49 & 0.55 & 0.59 & 0.40 & 0.48 & 0.66 & 0.67 \\
\midrule
MAJ & 0.66 & 0.56 & 0.72 & 0.50 & 0.70 & 0.55 & \textbf{0.83} & 0.64 & 0.76 \\
BAU & 0.62 & 0.61 & 0.64 & 0.52 & 0.58 & 0.44 & 0.61 & 0.62 & 0.71 \\
MAS & 0.65 & 0.41 & 0.64 & 0.59 & 0.65 & 0.59 & 0.76 & 0.60 & 0.72 \\
SAD & 0.65 & 0.59 & 0.57 & 0.54 & 0.61 & 0.53 & 0.71 & 0.58 & 0.71 \\
\bottomrule
\end{tabular}}
\label{tab:segment_f1}
\end{table}



\section{Baselines}
In this section, we describe basic experiments performed on the given dataset as baselines for future research\footnote{\url{https://github.com/mbzuai-nlp/CASA.git}}. 
We split the clips into 5-second segments with a 2-second overlap window. We aggregate the labels of the segments in two stages: first, we use the labels from the best aggregation method (MAJ) described in \ref{sec:dataprep}. For each segment, we then give a label of 1 for each type if \textit{any} instance of that type occurs within the segment and 0 otherwise, resulting in a `mutli-label' setup. We also add a class \texttt{Any},  which is 1 if any disfluency or secondary event exists in a segment; this accounts for overall disfluency detection performance. As we have different types of annotations, we experimented with speech-only models for disfluency types, vision-only models for secondary behaviors, and multi-modal model
for both disfluency and secondary behaviors.


\textbf{Speech-Based Disfluency Classification:}
We used the pre-trained \emph{wav2vec} \cite{wav2vec} model fine-tuned for audio classification tasks as an audio feature extractor, followed by 2 linear layers with RELU activation,  a classifier head and a Sigmoid activating layer for the probability distribution. The wav2vec model weights were used offline. We used weighted binary cross-entropy loss as our dataset is highly imbalanced. The input to wav2vec is the raw audio waveform sampled at 16kHz. 
\textbf{Secondary Event Classification:}
We used the extracted video segments described above for secondary event classification. 
We used the VIVIT \cite{vivit} model as the video feature encoder, followed by 2 linear layers and a classification head. We keep the VIVIT model frozen except for the embedding, first and last encoder layers, and it's classification head. The input to the video model is 10 frames of $224 \times 224$ images sampled from the input video.
\textbf{Multi-modal Stuttering Classification:}
Here, we attempt stuttering moment classification using information from both the speech and video segments. 
We use the same architecture above as our Audio feature encoder. We concatenate the normalised output from wav2vec encoder and VIVIT's classifier head (same configuration above). We follow the concatenation with a fully connected linear layer followed by GELU activation, another linear layer, and a classification head.

The results in table \ref{tab:result-total} show that the multi-modal approach is best for stuttering event detection (the `Any' class) and is generally better for secondary behaviors. For primary disfluency types, the baseline acoustic model achieved the best performance. Combining the training data for both reading and interview sections results in improved performance (the \textbf{Total} columns). Future work may experiment with hybrid approaches or more integrated multi-modal methods. 
\begin{table}[htp!]
\caption{F1 scores per type for baseline models. \textbf{Reading} and \textbf{Interview} are for models trained on these sections separately. \textbf{Total} is combined training. }
\centering
\small
\setlength{\tabcolsep}{4pt} 
\renewcommand{\arraystretch}{1} 
\scalebox{0.8}{
\begin{tabular}{@{}lccc|ccc|ccc@{}}
\toprule
 & \multicolumn{3}{c}{\textbf{Reading}} & \multicolumn{3}{c}{\textbf{Interview}}  & \multicolumn{3}{c}{\textbf{Total}}\\ 
\cmidrule(lr){2-4} \cmidrule(lr){5-7} \cmidrule(lr){8-10}
\textbf{Behavior} & \textbf{Aud} & \textbf{Vid} & \textbf{Both} & \textbf{Aud} & \textbf{Vid} & \textbf{Both} &\textbf{Aud} & 
\textbf{Vid} & \textbf{Both} \\ 
\midrule
SR   & 0.38 & -   & 0.07  & 0.45 & -   & 0.10 & \textbf{0.48 }& -   & 0.11\\ 
ISR  & 0.43  & -   & 0.58  & 0.53  & -   & 0.61 & 0.59  & -   & \textbf{0.62} \\ 
MUR  & 0.27 & -   & 0.24  & 0.29  & -   & 0.20 & \textbf{0.35}  & -   & 0.10 \\ 
P    & \textbf{0.22} & -   & 0.21  & 0.20  & -   & 0.20 & 0.20  & -   & 0.19 \\ 
B    & \textbf{0.68} & -   & 0.19  & 0.61  & -   & 0.30 & \textbf{0.68}  & -   & 0.42 \\ 
V    & -   & 0.18 & 0.15  & -   & 0.20  & 0.19& -   & 0.21  & \textbf{0.23} \\ 
FG   & -   & 0.57  & 0.66  & -   & 0.61  & 0.69 & -   & 0.63  & \textbf{0.72} \\ 
HM   & -   & 0.27  & 0.54  & -   & 0.25  & 0.52 & -   & 0.28  & \textbf{0.57} \\ 
Any  & 0.90  & 0.92  & 0.92  & 0.92   & 0.90  & 0.94 & 0.93   & 0.92  & \textbf{0.95} \\ 
\bottomrule
\end{tabular}
}
\label{tab:result-total}
\end{table}


\section{Conclusion}

We presented a clinically annotated dataset for stuttering severity assessment. Our analysis and baseline results illustrate the complexity of the task that necessitates further investigations. The annotations and baseline scripts will be released to encourage researchers to explore this clinical application beyond standard disfluency detection and classification.



\bibliographystyle{IEEEtran}
\bibliography{mybib}

\end{document}